\documentclass[sigconf]{acmart}

\usepackage{amsmath}
\usepackage{booktabs}
\usepackage{graphicx}
\usepackage{tikz}
\usepackage{enumitem}
\usepackage{bm}
\usepackage{makecell}
\usepackage{subcaption}
\usetikzlibrary{arrows.meta,positioning,fit,calc}

\usepackage{siunitx}

\setcopyright{none}
\copyrightyear{2027}
\acmYear{2027}
\acmDOI{}
\acmConference[WWW '27]{Proceedings of the ACM Web Conference 2027}{TBD}{TBD}
\acmISBN{}
\renewcommand\footnotetextcopyrightpermission[1]{}

\begin{document}

\title{EASformer: Environment-Action-State Alignment for \\Industrial Generative Ranking}

\author{Bohao Li}
\authornote{The first four authors contributed equally to this work.}
\email{libh@buaa.edu.cn}
\affiliation{%
  \institution{Beihang University}
  \city{Beijing}
  \country{China}}

\author{Dinglun Cao}
\authornotemark[1]
\email{cdinglun@buaa.edu.cn}
\affiliation{%
  \institution{Beihang University}
  \city{Beijing}
  \country{China}}

\author{Pengcheng Huang}
\authornotemark[1]
\email{24s151025@stu.hit.edu.cn}
\affiliation{%
  \institution{Harbin Institute of Technology}
  \city{Shenzhen}
  \country{China}}

\author{Yingda Tang}
\authornotemark[1]
\email{ttttang2002@gmail.com}
\affiliation{%
  \institution{Independent Researcher}
  \city{Shanghai}
  \country{China}}

\author{Zhiting Zhao}
\email{24S151185@stu.hit.edu.cn}
\affiliation{%
  \institution{Harbin Institute of Technology}
  \city{Shenzhen}
  \country{China}}

\author{Yi Cao}
\authornote{Corresponding authors: Yi Cao and Bowen Du.}
\email{ycaoxd@163.com}
\affiliation{%
  \institution{Independent Researcher}
  \city{Shanghai}
  \country{China}}

\author{Jilei Fu}
\email{grayfu@gmail.com}
\affiliation{%
  \institution{Independent Researcher}
  \city{Shanghai}
  \country{China}}

\author{Heng Chang}
\email{heng.chang@huawei.com}
\affiliation{%
  \institution{Independent Researcher}
  \city{Beijing}
  \country{China}}

\author{Tianying Liu}
\email{tianying\_{}liu@outlook.com}
\affiliation{%
  \institution{Independent Researcher}
  \city{Shanghai}
  \country{China}}

\author{Miao Yan}
\email{erin.mia.yan@gmail.com}
\affiliation{%
  \institution{Independent Researcher}
  \city{Shanghai}
  \country{China}}

\author{Zhende Zhuang}
\email{zhuangjz4@gmail.com}
\affiliation{%
  \institution{Independent Researcher}
  \city{Shanghai}
  \country{China}}

\author{Jiarui Wang}
\email{2296339656@qq.com}
\affiliation{%
  \institution{Independent Researcher}
  \city{Shanghai}
  \country{China}}

\author{Deliang Liu}
\email{deliang\_{}liu1995@163.com}
\affiliation{%
  \institution{Independent Researcher}
  \city{Shanghai}
  \country{China}}

\author{Bowen Du}
\authornotemark[2]
\email{dubowen@buaa.edu.cn}
\affiliation{%
  \institution{Beihang University}
  \city{Beijing}
  \country{China}}

\begin{abstract}
Modern industrial ranking systems increasingly adopt generative recommendation (GR) paradigms, in which user behavior sequences and non-sequential (NS) features are tokenized and jointly modeled by Transformer-style backbones. This unified formulation provides a scalable interface for modeling long histories and heterogeneous features. 
However, sequence and NS representations often carry asymmetric semantics: historical representations may combine environment context with observed-action semantics, whereas candidate-side NS representations contain no observed action but incorporate richer supplementary context, such as user-item interaction features and recent behavior lists. 
Direct interaction between such representations can make query-key dot products compare different semantic objects, limiting meaningful cross-token interaction. 
To fill this gap, we propose \textbf{EASformer}, an \textbf{E}nvironment-\textbf{A}ction-\textbf{S}tate disentangled architecture for industrial app ranking.
EASformer constructs an explicitly aligned context environment space shared by historical and non-sequence tokens, while disentangling user actions from historical token and candidate-specific context form non-sequence tokens. Based on this aligned space, EASformer enables complementary historical action retrieval and state-conditioned action prediction. Furthermore, the disentangled design supports a Cloze-style self-supervised objective by masking historical actions and reusing the candidate prediction pathway. On an offline benchmark from a real-world industrial app marketplace, EASformer achieves an AUC of 0.89411, yielding a 0.88\% relative improvement over the deployed baseline. 
Online A/B tests further support this advantage, with relative eCPM lifts of 1.26\% and 3.94\% across two real-world ranking scenarios.
\end{abstract}



\maketitle

\section{Introduction}

\begin{figure}[t]
  \centering
  \includegraphics[width=0.9\linewidth]{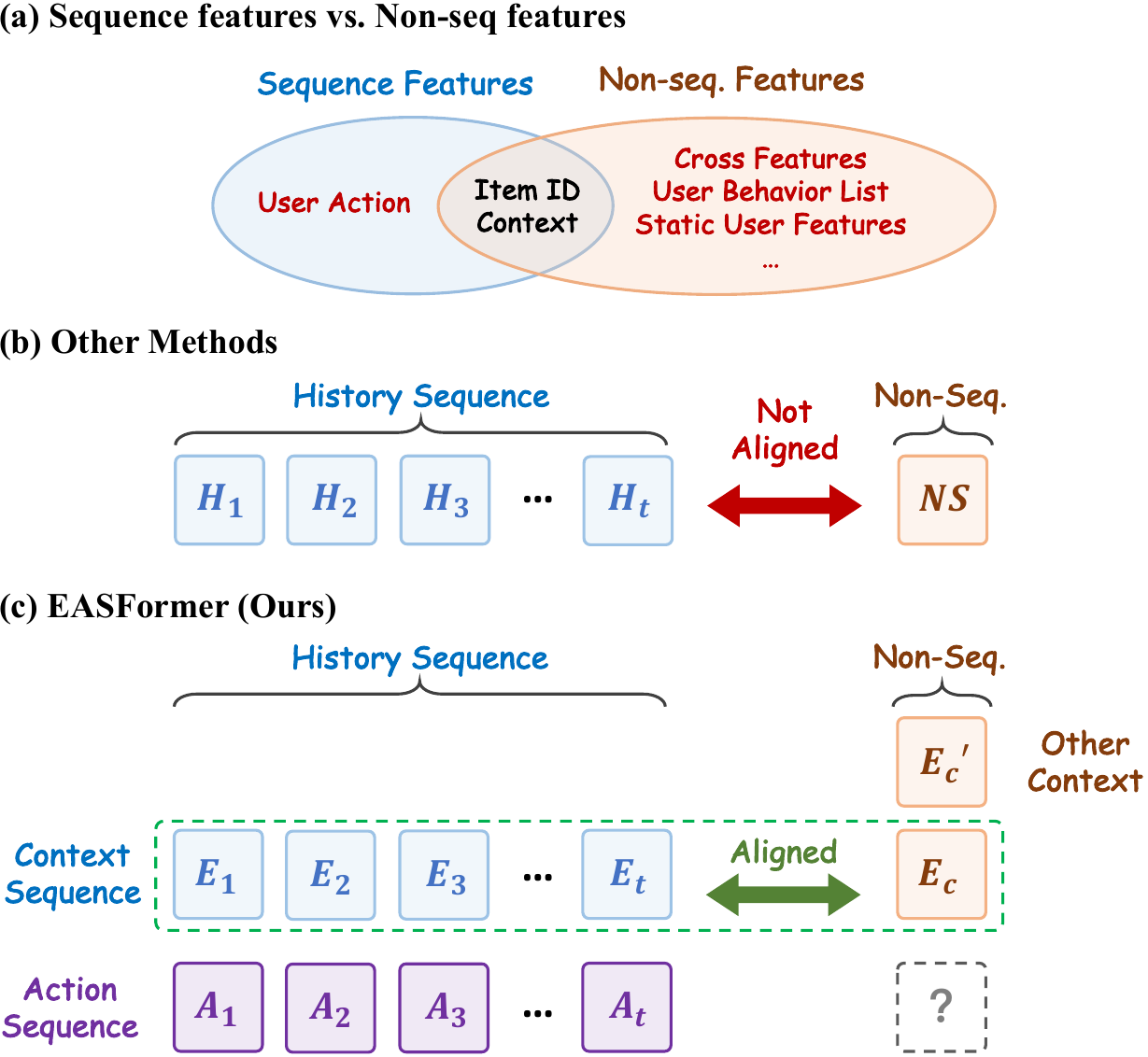}
  \caption{Semantic misalignment between sequence and NS representations. EASformer separates aligned context environment \(E\), candidate-specific supplementary context \(E'_c\), and user action before interaction.}
  \Description{A schematic comparison between existing tokenization and EASformer. Existing methods mix sequence and non-sequential features with unequal semantics, while EASformer aligns historical context tokens with candidate context tokens and separates actions as values.}
  \label{fig:token-misalignment}
\end{figure}

Modern industrial recommender systems are shifting from conventional feature-engineered ranking models toward generative recommendation (GR) paradigms~\cite{zhai2024hstu,deng2025onerec,hou2025actionpiece}. Unlike traditional architectures built on manually designed feature interactions, GR models represent diverse information sources as tokens and process them with Transformer-based backbones. This unified formulation provides a scalable interface for jointly modeling long user behavior sequences, candidate information, and heterogeneous side features, making it increasingly attractive for large-scale industrial ranking.

Recent studies have advanced token-based recommendation along several directions. Sequential methods, such as SASRec and BERT4Rec, model user behaviors as ordered token sequences~\cite{kang2018sasrec,sun2019bert4rec}, while later works enrich sequence representations with user actions or item-level semantics~\cite{xia2023transact,hou2025actionpiece,sun2026beyonditemids}. HSTU reformulates industrial ranking as generative sequential transduction, while MTGR adapts its backbone with candidate-side cross features and discriminative training~\cite{zhai2024hstu,han2025mtgr}. In parallel, heterogeneous feature interaction has been explored through self-attention, co-action modeling, and token mixing~\cite{song2018autoint,bian2020can,zhu2025rankmixer}. Industrial rankers also integrate sequential behaviors and non-sequential (NS) features through unified or modular architectures, such as OneTrans, InterFormer, HyFormer, MixFormer, HeMix, and EST~\cite{zhang2025onetrans,zeng2025interformer,huang2026hyformer,huang2026mixformer,wang2026hemix,liu2026est}. Consequently, integrating sequence and NS tokens has become a central research problem in token-based industrial ranking.

However, existing token-based GR methods mainly emphasize token expressiveness and fusion capacity, while overlooking a fundamental requirement for token interaction: \textbf{semantic alignment}. Query-key similarity in self-attention is meaningful only when the compared representations are semantically comparable~\cite{vaswani2017attention}. This condition naturally holds for linguistic tokens in NLP~\cite{devlin2019bert}, visual patches in computer vision~\cite{dosovitskiy2021vit}, and explicitly aligned representations in multimodal learning~\cite{radford2021clip}. In industrial GR, however, heterogeneous semantics may be combined during token construction or entangled through subsequent model interactions. The resulting tokens therefore lack a consistent semantic basis for comparison, making their query-key dot products difficult to interpret as semantic similarity.


Taking app recommendation as an example, a ranking model must jointly leverage historical app interactions, candidate attributes, user-item interaction features, and contextual signals such as device, time, and location. As illustrated in Figure~\ref{fig:token-misalignment}, a historical interaction comprises an environment context and an observed user action. In contrast, the candidate-side NS representation contains the corresponding current context environment but no observed action, while also incorporating candidate-specific supplementary context, such as user-item cross features, recent behavior lists, and user attributes. Sequence and NS representations therefore have asymmetric semantic compositions, with each containing information absent from the other. Directly interacting such representations can cause query-key dot products to compare different semantic objects, thereby limiting the ability of Transformer-based architectures to model meaningful cross-token interactions.

Motivated by this limitation, we seek to disentangle heterogeneous semantics before token interaction while preserving a shared embedding space for aligned fields. Specifically, each historical interaction is separated into an aligned environment representation and an observed user action, while candidate-side information is divided into the corresponding current environment and candidate-specific supplementary context. Historical and candidate environments are constructed from semantically matched fields and embedded in a shared environment space.

Building on this alignment-and-disentanglement design, we propose \textbf{EASformer} for industrial GR ranking. EASformer models user responses from two complementary views. The \emph{temporal retrieval view} matches the current environment with historical environments and retrieves their associated actions as personalized evidence. The \emph{spatial mapping view} models a state-conditioned context-to-action relationship from the current environment and routed user state. The two views are complementary: temporal retrieval uses concrete historical evidence, while spatial mapping captures context-dependent behavior beyond direct historical matches. Their outputs are adaptively fused, and supplementary context is incorporated through a separate pathway for final ranking.
The explicit separation of actions further enables a natural Cloze-style masked-action objective. During training, selected historical actions are masked and reconstructed through the same dual-view pathway and ranking head. This objective leverages historical environment-to-action relationships as additional supervision for both prediction views.
We evaluate EASformer on an industrial app-ranking task. On a 25-day offline benchmark, EASformer achieves an AUC of 0.89411, corresponding to a 0.88\% relative improvement over the production baseline (HyFormer). 
Further online A/B tests of the lightweight EASformer demonstrate consistent gains in two real-world industrial ranking scenarios.

\begin{figure*}[t]
    \centering
    \includegraphics[width=1.0\linewidth]{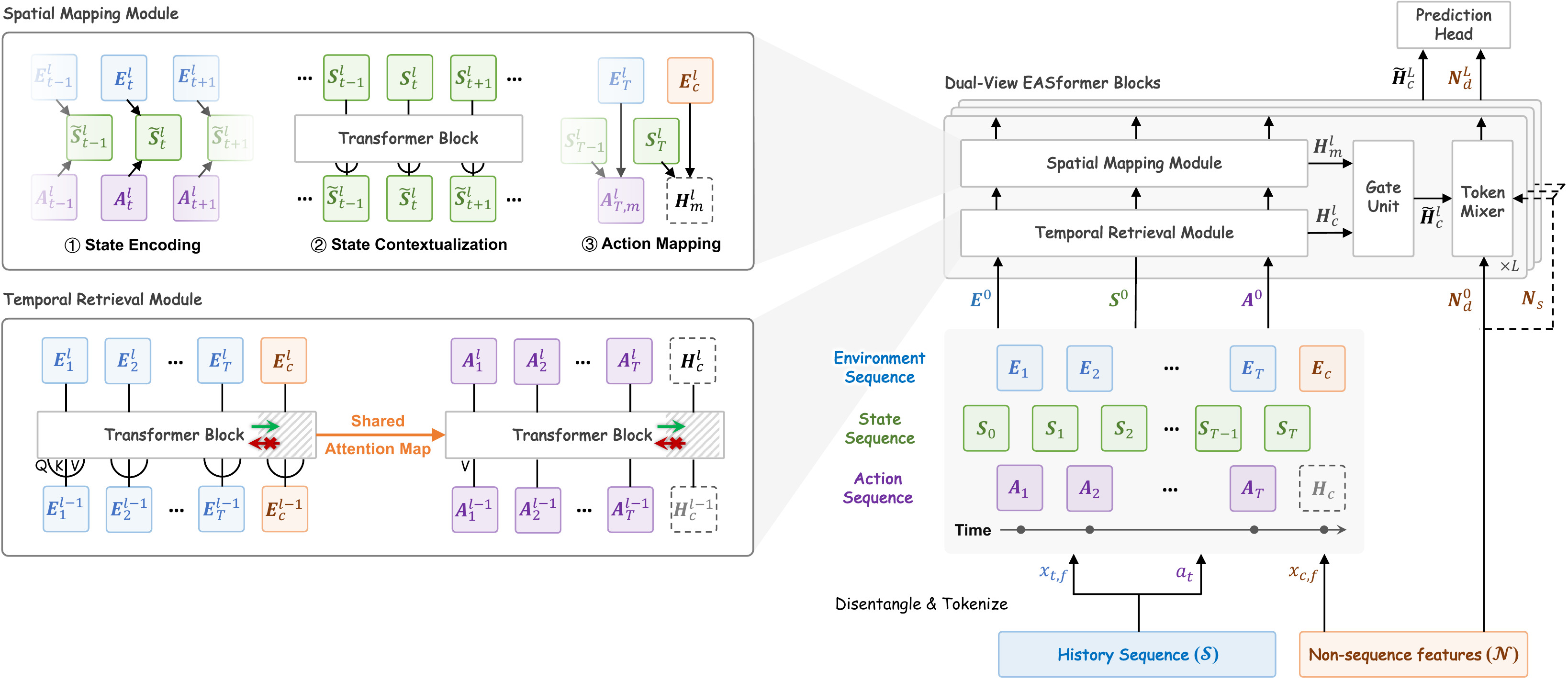}
    \caption{Overview of the proposed framework.}
    \label{fig:framework}
\end{figure*}

Our main contributions are summarized as follows:
\begin{itemize}[
    leftmargin=*,
    topsep=2pt,
    itemsep=2pt,
    parsep=0pt,
    partopsep=0pt
]
  \item We identify semantic misalignment between sequence and NS representations as a practical limitation in industrial GR ranking and formulate aligned environment modeling as a design principle for their interaction.

  \item We propose \textbf{EASformer}, an Environment-Action-State disentangled architecture that constructs an aligned environment space from semantically matched historical and candidate fields. EASformer combines temporal retrieval of historical action evidence with spatial, state-conditioned context-to-action mapping for candidate prediction.

  \item We introduce a Cloze-style masked-action objective that reuses the dual-view candidate prediction pathway and leverages logged environment-to-action relationships as additional supervision.

  \item We evaluate EASformer in real-world industrial app-ranking settings, achieving an AUC of 0.89411 and a 0.88\% relative improvement over the deployed baseline offline, while the lightweight version of EASformer yields relative eCPM lifts of 1.26\% and 3.94\% in two online A/B tests.
\end{itemize}

\section{Preliminary: Industrial Generative Ranking}
\label{sec:preliminaries}

For each ranking request of user $u$, let
\begin{equation}
\mathcal{S}_{u}
=
\left(
\bm{x}_{1},
\bm{x}_{2},
\ldots,
\bm{x}_{T}
\right),
\end{equation}
denote the sequential features, where $\mathcal{S}_{u}$ is an ordered
sequence of $T$ historical interactions. Each interaction
$\bm{x}_{t}$ contains its contextual information, such as item and
scenario attributes, together with the observed user action.

Given a recalled candidate set $\mathcal{C}_{u}$, each candidate
$c\in\mathcal{C}_{u}$ is associated with non-sequential features
$\mathcal{N}_{u,c}$. These features include candidate attributes,
user static information, user-item cross statistics, and
auxiliary behavior lists.

Industrial generative ranking jointly models $\mathcal{S}_{u}$ and
$\mathcal{N}_{u,c}$ to estimate the user's response to each candidate:
\begin{equation}
\hat{y}_{u,c}
=
P\!\left(
y_{u,c}=1
\mid
\mathcal{S}_{u},
\mathcal{N}_{u,c}
\right),
\qquad
c\in\mathcal{C}_{u},
\label{eq:industrial_ranking}
\end{equation}
where $y_{u,c}\in\{0,1\}$ is the supervision label and
$\hat{y}_{u,c}\in(0,1)$ is the predicted ranking score. The candidates
are ranked according to their predicted scores.
As candidates are modeled independently, we focus on a single candidate and, for simplicity, omit the subscripts hereafter, using $\mathcal{S}$, $\mathcal{N}$, $y$, and $\hat{y}$.
\section{Method}

EASformer follows a disentanglement-and-alignment principle, aiming to construct an aligned context-environment space for measuring the relevance between historical interactions and the current candidate. As illustrated in Figure~\ref{fig:framework}, it extracts semantically matched context-environment fields from the sequential and candidate-side non-sequential (NS) inputs, while separately encoding observed actions and the complete candidate-side NS tokens. EASformer then progressively refines a candidate-conditioned user-action representation through $L$ stacked blocks, each comprising temporal retrieval, spatial mapping, and heterogeneous information fusion. The disentangled Action stream further enables a Cloze-style masked-action objective that reuses the same prediction pathway to provide auxiliary supervision.

\subsection{Aligned and Disentangled Tokenization}
\label{sec:tokenization}

We assume that history--candidate relevance should be measured through
context-environment fields with aligned semantics. Let
$\mathcal{F}_{E}$ denote the set of such fields shared by the
historical interactions and the current candidate.

\textit{Context Environment.}
For each historical interaction, we extract its context-environment
fields to construct an Environment token. The corresponding fields are
also extracted from the candidate-side non-sequential input to construct
the current Environment token:
\begin{equation}
\bm{E}^{0}_{i}
=
f_E\!\left(
[\phi_f(x_{i,f})]_{f\in\mathcal{F}_{E}}
\right),
\qquad
i\in\{1,\ldots,T,c\},
\label{eq:environment_tokenization}
\end{equation}
where $\phi_f(\cdot)$ is the field-specific transformation, brackets
denote concatenation, and $f_E$ is shared across the historical and
candidate sides. The historical Environment sequence is denoted by
$\bm{E}^{0}=[\bm{E}^{0}_{1},\ldots,\bm{E}^{0}_{T},\bm{E}^{0}_{c}]$, while
$\bm{E}^{0}_{c}$ denotes the context of the candidate.

\textit{Action.}
After separating the environment fields, we independently encode the observed user action $a_t$ of each interaction as
\begin{equation}
\bm{A}^{0}_{t}=e_A(a_t),
\qquad
t=1,\ldots,T,
\label{eq:action_tokenization}
\end{equation}
where $e_A(\cdot)$ is a learnable action embedding table. Since the action associated with the current candidate is unobserved and is the target to be inferred, we initialize it with a learnable placeholder,
$\bm{H}^{0}_{c}=\bm{h}_{\mathrm{unk}}$.
The resulting Action sequence is $\bm{A}^{0} = [\bm{A}^{0}_{1},\ldots,\bm{A}^{0}_{T},\bm{H}^{0}_{c}]$.

\textit{User State.}
To model how user preferences evolve along the interaction sequence, we
introduce a temporally staggered State sequence
$\bm{S}^{0}=[\bm{S}^{0}_{0},\ldots,\bm{S}^{0}_{T}]$, 
as illustrated in
Figure~\ref{fig:framework}. All state tokens are initialized
by the same embedding of static user features, such as device
attributes:
\begin{equation}
\bm{S}^{0}_{t}=f_U(\mathcal{N}_{U}),
\quad
t=0,\ldots,T,
\label{eq:state_initialization}
\end{equation}
where $\mathcal{N}_{U}\subset\mathcal{N}$ denotes the user-static
feature subset.

\textit{Non-Sequential Information.}
Although the aligned candidate fields are extracted to construct
$\bm{E}^{0}_{c}$, the complete non-sequential input remains useful for
ranking. We therefore independently map the full $\mathcal{N}$ into $K$ static NS tokens:
\begin{equation}
\bm{N}_{s}
=
f_N(\mathcal{N})
=
[\bm{N}_{s,1},\ldots,\bm{N}_{s,K}]\in
\mathbb{R}^{K\times \mathcal{D}_{\mathrm{NS}}},
\qquad
\bm{N}^{0}_{d}=\bm{N}_{s},
\label{eq:ns_tokenization}
\end{equation}
where, $\bm{N}_{s}$ remains fixed across all blocks, whereas $\bm{N}^{\ell}_{d}$ denotes the dynamically updated NS tokens after the $\ell$-th block. 
The Environment, Action and State representations lie in $\mathbb{R}^\mathcal{D}$, while each NS token lies in $\mathbb{R}^{\mathcal{D}_{\mathrm{NS}}}$.

\subsection{Dual-View EASformer Block}
\label{sec:EASformer_block}

EASformer stacks $L$ blocks to progressively update the Environment, Action, State, and dynamic NS representations. Each block consists of a Temporal Retrieval Module, a Spatial Mapping Module, and a heterogeneous information fusion module. The first two modules produce complementary candidate-action representations, which are subsequently fused to interact with the NS tokens.

\subsubsection{\textbf{Temporal Retrieval.}}
We assume that actions under similar context environments provide direct evidence for predicting the current candidate action. Therefore, the Temporal Retrieval Module computes attention solely from the Environment sequence and shares the resulting attention map with the Action sequence.

At the $\ell$-th block, let $\bm{E}^{\ell-1}$ and $\bm{A}^{\ell-1}$ denote the complete Environment and Action sequences, whose candidate slots are $\bm{E}^{\ell-1}_{c}$ and $\bm{H}^{\ell-1}_{c}$, respectively. The shared attention map is
\begin{equation}
\bm{P}^{\ell}
=
\operatorname{Softmax}\!\left(
\frac{
(\bm{E}^{\ell-1}\bm{W}^{\ell}_{Q})
(\bm{E}^{\ell-1}\bm{W}^{\ell}_{K})^{\top}
}{
\sqrt{d_h}
}
+
\bm{M}
\right),
\label{eq:shared_attention}
\end{equation}
where $d_h$ is the head dimension, $\bm{W}^{\ell}_{Q}$ and $\bm{W}^{\ell}_{K}$ are learnable projection matrices, and $\bm{M}$ is an asymmetric visibility mask. Historical tokens attend only to valid historical positions and cannot access the candidate slot, whereas the candidate token attends to itself and all eligible preceding interactions. Then, the Environment and Action streams use separate value
projections but share $\bm{P}^{\ell}$:
\begin{equation}
\bm{\widetilde{E}}^{\ell}
=
\bm{P}^{\ell}
\bm{E}^{\ell-1}\bm{W}^{\ell}_{V_E},
\qquad
\bm{\widetilde{A}}^{\ell}
=
\bm{P}^{\ell}
\bm{A}^{\ell-1}\bm{W}^{\ell}_{V_A},
\label{eq:shared_attention_values}
\end{equation}
where $\bm{W}^{\ell}_{V_E}$ and $\bm{W}^{\ell}_{V_A}$ are independent value projection matrices. The two streams are then independently updated through residual attention and feed-forward sublayers:
\begin{equation}
\begin{aligned}
\widehat{\bm{E}}^{\ell}
&=
\operatorname{RN}\!\left(
\bm{E}^{\ell-1}
+
\widetilde{\bm{E}}^{\ell}
\right),
\qquad
\bm{E}^{\ell}
=
\operatorname{RN}\!\left(
\widehat{\bm{E}}^{\ell}
+
\operatorname{FFN}^{\ell}_{E}
\left(
\widehat{\bm{E}}^{\ell}\right)
\right),
\\
\widehat{\bm{A}}^{\ell}
&=
\operatorname{RN}\!\left(
\bm{A}^{\ell-1}
+
\widetilde{\bm{A}}^{\ell}
\right),
\qquad
\bm{A}^{\ell}=
\operatorname{RN}\!\left(
\widehat{\bm{A}}^{\ell}
+
\operatorname{FFN}^{\ell}_{A}
\left(\widehat{\bm{A}}^{\ell}\right)
\right).
\end{aligned}
\label{eq:temporal_update}
\end{equation}
Here, $\text{RN}(\cdot)$ denotes RMS normalization, $\operatorname{FFN}^{\ell}_{E}(\cdot)$ and $\operatorname{FFN}^{\ell}_{A}(\cdot)$ are the feed-forward networks \cite{vaswani2017attention}, and different occurrences use independent parameters. 

The candidate slot $\bm{H}^{\ell}_{c} \in \bm{A}^{\ell}$  serves as the output of the temporal retrieval module. In this way, the Environment stream determines where to retrieve, while the Action stream provides the corresponding behavioral evidence.

\subsubsection{\textbf{Spatial Mapping.}}

Temporal retrieval provides explicit behavioral evidence from similar historical environments, but it relies on the assumption that similar contexts lead to consistent user actions. However, user behavior can evolve over time as the underlying user state changes. Therefore, EASformer introduces a spatial mapping module to learn a state-conditioned action representation, which captures how the current context leads to different action tendencies under different user states.

\textit{State Encoding.}
The State sequence captures the evolving user behavioral state by aggregating information from historical interactions.  At the $\ell$-th block, each historical Environment-Action pair is first projected into the corresponding state position:
\begin{equation}
\widetilde{\bm{S}}^{\ell}_{t}
=
\operatorname{RN}
\left(
\bm{S}^{\ell-1}_{t}
+
f^{\ell}_{EA}
(
[\bm{E}^{\ell}_{t};
\bm{A}^{\ell}_{t}]
)
\right),
\qquad
t=0,\ldots,T,
\label{eq:state_encoding}
\end{equation}
where $f^{\ell}_{EA}(\cdot)$ is a learnable mapping function that encodes the interaction-level context and action information into the user state.

\textit{State Contextualization.}
Since user behavioral states evolve across interactions, the encoded state sequence is further contextualized by a Transformer encoder block:
\begin{equation}
\bm{S}^{\ell}
=
\operatorname{Transformer}^{\ell}_{S}
(
\widetilde{\bm{S}}^{\ell}
),
\label{eq:state_contextualization}
\end{equation}
where each state token captures dependencies among historical states.
The latest state token after contextualization, denoted as $\bm{S}^{\ell}_{T} \in \bm{S}^{\ell}$, summarizes the user's current behavioral state.

\textit{Action Mapping.}
The spatial mapping module then predicts the candidate action tendency conditioned on the current context environment and routed user state:
\begin{equation}
\bm{H}^{\ell}_{m}
=
\operatorname{RN}
\left(
\bm{H}^{\ell-1}_{m}
+
f^{\ell}_{M}
(
[\bm{E}^{\ell}_{c};
\bm{S}^{\ell}_{T}]
)
\right),
\label{eq:action_mapping}
\end{equation}
where $f^{\ell}_{M}(\cdot)$ is a learnable mapping function. The resulting $\bm{H}^{\ell}_{m}$ captures how the current context and user state jointly determine the candidate action beyond direct historical retrieval.



\subsubsection{\textbf{Dual-View Fusion.}}

The Temporal Retrieval and Spatial Mapping modules capture candidate
actions from complementary perspectives. The temporal view provides
explicit behavioral evidence retrieved from similar historical
environments, while the spatial view models state-conditioned action
tendencies.

Before fusion, the two representations are projected into the NS token space via $\overline{\bm{H}}^{\ell}_{c}
=
\pi^{\ell}_{T}
\left(
\bm{H}^{\ell}_{c}
\right)$ and $\overline{\bm{H}}^{\ell}_{m}
=
\pi^{\ell}_{M}
\left(
\bm{H}^{\ell}_{m}
\right)$,
where $\pi^{\ell}_{T},\pi^{\ell}_{M}:\mathbb{R}^{\mathcal{D}}\rightarrow\mathbb{R}^{\mathcal{D}_{\mathrm{NS}}}$ are learnable linear projections. When $\mathcal{D}=\mathcal{D}_{\mathrm{NS}}$, they can be implemented as identity mappings.

The projected representations are then combined through a vector-wise gating mechanism:
\begin{equation}
\begin{aligned}
\bm{\gamma}^{\ell}
&=
\sigma
\left(
\bm{W}^{\ell}_{g}
[
\overline{\bm{H}}^{\ell}_{m};
\overline{\bm{H}}^{\ell}_{c}
]
+
\bm{b}^{\ell}_{g}
\right),
\\
\widetilde{\bm{H}}^{\ell}_{c}
&=
\operatorname{RN}
\left(
\bm{\gamma}^{\ell}
\odot
\overline{\bm{H}}^{\ell}_{m}
+
(1-\bm{\gamma}^{\ell})
\odot
\overline{\bm{H}}^{\ell}_{c}
\right).
\end{aligned}
\label{equ:gate}
\end{equation}

The fused representation $\widetilde{\bm{H}}^{\ell}_{c}
\in\mathbb{R}^{D_{\mathrm{NS}}}$ adaptively combines state-conditioned action mapping and retrieved historical evidence for the current candidate.

\subsubsection{\textbf{Heterogeneous Token Interaction.}}

Candidate-side NS features provide additional rich serving-time information, but they are not included in the aligned environment space. Therefore, EASformer further incorporates the NS token.

Specifically, the dynamic NS tokens are updated by interacting with the current candidate representation and static NS tokens:
\begin{equation}
\bm{N}^{\ell}_{d}
=
\operatorname{TokenMixer}^{\ell}
\left(
[
\bm{N}^{\ell-1}_{d};
\widetilde{\bm{H}}^{\ell}_{c};
\bm{N}_{s}
]
\right),
\end{equation}
where $\bm{N}_{s}$ denotes static NS tokens generated during tokenization and $\bm{N}^{\ell}_{d}$ denotes dynamically updated NS representations. $\operatorname{TokenMixer}(\cdot)$ \cite{tolstikhin2021mlp, zhu2025rankmixer} performs token-level interaction among heterogeneous NS representations and candidate state. 

This design allows rich candidate-side information to enhance final ranking while preserving the semantic purity of the aligned context-environment space.

\subsection{Prediction Head and Main Objective}
\label{sec:prediction}

After $L$ EASformer blocks, the final candidate representation and dynamic NS tokens are concatenated and fed into a residual MLP prediction head:
\begin{equation}
\hat{y}
=
\sigma\!\left(
f_{P}\left(\left[
\widetilde{\bm{H}}^{L}_{c};
\operatorname{vec}(\bm{N}^{L}_{d})
\right]\right)
\right),
\label{eq:main_prediction}
\end{equation}
where $\operatorname{vec}(\cdot)$ concatenates the dynamic NS tokens, $f_P(\cdot)$ denotes the prediction head, and $\sigma(\cdot)$ is the sigmoid function. The main ranking objective is
\begin{equation}
\mathcal{L}_{\mathrm{rank}}
=
\operatorname{BCE}(y,\hat{y}),
\label{eq:ranking_loss}
\end{equation}
which is averaged over valid candidates during training.

\begin{figure}[t]
    \centering
    \includegraphics[width=1.0\linewidth]{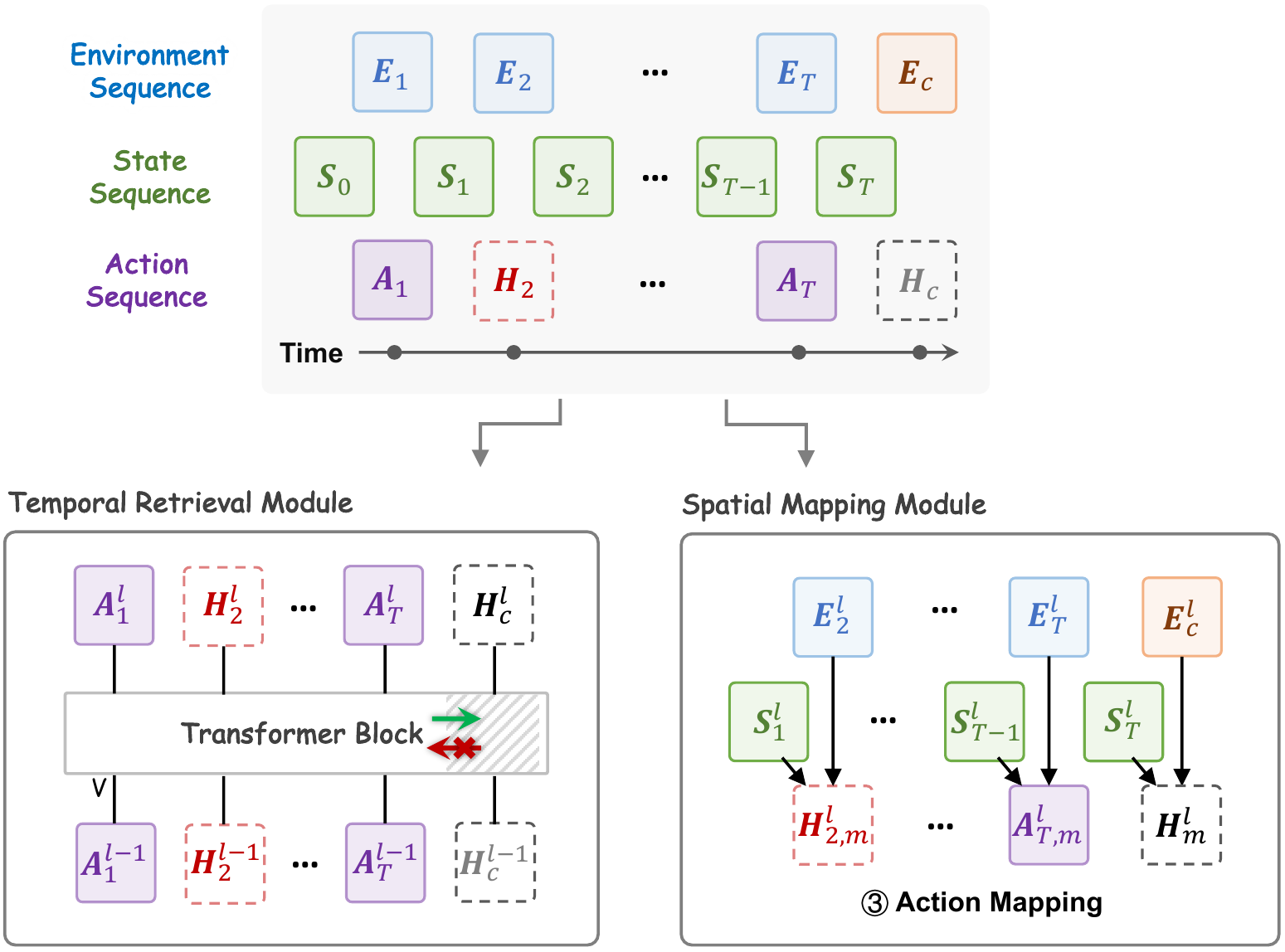}
    \caption{Masked-action self-supervision in EASformer. Masked historical Action tokens are replaced with a learnable placeholder and processed normally by both the Temporal Retrieval and Spatial Mapping modules. 
    }
    \label{fig:self_supervision}
\end{figure}

\subsection{Masked-Action Self-Supervision}
\label{sec:self_supervision}

The disentangled Action stream provides a natural way to exploit the environment-to-action relations contained in historical interactions. Specifically, EASformer masks selected historical actions while retaining their corresponding Context Environment tokens, thereby turning these positions into pseudo-candidates with known environments but unknown actions.

Let $\rho$ denote the masking ratio, and let
$\mathcal{M}\subseteq\{1,\ldots,T\}$ be the sampled positions, where
$|\mathcal{M}|\approx\rho T$. For each $t\in\mathcal{M}$, the original
Action token is replaced with the same learnable placeholder used for
the candidate:
\begin{equation}
\bm{A}^{0}_{t}
=
\bm{H}^{0}_{t}
=
\bm{h}_{\mathrm{unk}},
\qquad t\in\mathcal{M}.
\label{eq:masked_action}
\end{equation}

The masked position is then processed by the same $L$ EASformer blocks.
At each block $\ell$, the masked Action sequence is processed by the same backbone as the original Action sequence: 1) In the Temporal Retrieval Module, the attention maps derived from the Environment sequence update the masked action slot, producing the temporal representation $\bm{H}^{\ell}_{t}$. 2) In the Spatial Mapping Module, the masked Action token also participates in the state encoding, while the corresponding Environment token $\bm{E}^{\ell}_{t}$ and its latest State token $\bm{S}^{\ell}_{t-1}$ are mapped to the spatial representation $\bm{H}^{\ell}_{t,m}$. 
Since the observed action has been replaced by $\bm{h}_{\mathrm{unk}}$, these computations do not expose the target action. 

Across the $L$ blocks, temporal and spatial evidence is accumulated through residual updates, yielding $\bm{H}^{L}_{t}$ and $\bm{H}^{L}_{t,m}$, respectively. The two final representations are then fused using the same dual-view gate defined in Eq.~\ref{equ:gate}:
\begin{equation}
\widetilde{\bm{H}}^{L}_{t}
=
\operatorname{Gate}^{L}
\left(
\bm{H}^{L}_{t},
\bm{H}^{L}_{t,m}
\right).
\label{eq:aux_fusion}
\end{equation}

Since historical positions do not contain complete candidate-side NS features, we concatenate $\widetilde{\bm{H}}^{L}_{t}$ with $K$ learnable NS placeholder tokens $\bm{N}_{\mathrm{ph}} \in \mathbb{R}^{K\times \mathcal{D}}$ and reuse the main prediction head, $f_P(\cdot)$:
\begin{equation}
\hat{y}^{\mathrm{aux}}_{t}
=
\sigma\!\left(
f_P
\left(
[
\widetilde{\bm{H}}^{L}_{t};
\operatorname{vec}(\bm{N}_{\mathrm{ph}})
]
\right)
\right).
\label{eq:aux_prediction}
\end{equation}

The overall training objective is
\begin{equation}
\mathcal{L}
=
\mathcal{L}_{\mathrm{task}}
+
\frac{\lambda_s}{|\mathcal{M}|}
\sum_{t\in\mathcal{M}}
\operatorname{BCE}
\left(
a_{t},
\hat{y}^{\mathrm{aux}}_{t}
\right),
\label{eq:total_loss}
\end{equation}
where $\lambda_s$ denotes the weight of the self-supervised loss and $a_{t}$ is the original action at position $t$.
\section{Experiments}

We evaluate \textbf{EASformer} under a real-world industrial app-ranking benchmark and compare it with recent baselines. 
We further present ablation studies, case studies, parameter sensitivity analyses, and online A/B tests to further examine our method.

\subsection{Experimental Setup}

Here, we describe the dataset, evaluation protocol, baselines, and implementation details.

\subsubsection{\textbf{Industrial App-Ranking Dataset}}

We conduct experiments on a large-scale anonymized dataset collected from a real-world industrial app marketplace. Unlike conventional request-level ranking datasets, the data are organized at the user level, preserving each user's ordered interaction history across requests. The dataset is partitioned by day, with each partition containing interactions generated on that day, and spans more than 150 daily partitions.

To reduce data redundancy and training cost, negative exposure records are randomly sampled at a rate of 0.2. All user information is anonymized to protect privacy. 

\subsubsection{\textbf{Evaluation Protocol}}

We adopt a two-stage evaluation protocol. We first compare all models offline under identical data and input settings. After confirming the offline advantage of EAS for me r, we further conduct online A/B testing to evaluate its effectiveness in the production environment.

For offline evaluation, all models use the same temporal split: 25 consecutive daily partitions for training and the immediately following partition for evaluation. 
The Area Under the ROC Curve (AUC) is used as the primary offline evaluation metric.

For online A/B testing, EASformer is compared only against HyFormer, the incumbent production model, due to practical deployment and traffic constraints. 
Online performance is evaluated using eCPM (effective cost per mille), a monetization metric normalized per thousand impressions. Due to commercial confidentiality, we report only the relative eCPM lift over the control group.

\subsubsection{\textbf{Baselines}}


We select representative baseline methods and group them into three categories according to their primary modeling focus:
\begin{itemize}[
    leftmargin=*,
    topsep=2pt,
    itemsep=2pt,
    parsep=0pt,
    partopsep=0pt
]
    \item \textit{Feature interaction models.}
DCNv2~\cite{wang2021dcnv2}, RankMixer~\cite{zhu2025rankmixer}, and Wukong~\cite{wang2024wukong} focus on learning interactions among feature representations to capture cross-feature dependencies for ranking.

    \item \textit{Sequential models.}
SASRec~\cite{kang2018sasrec} and LONGER~\cite{chai2025longer} model dependencies within user behavior histories to extract interest representations for candidate prediction.

    \item \textit{Unified architectures.}
MTGR~\cite{han2025mtgr}, HyFormer~\cite{huang2026hyformer}, OneTrans~\cite{huang2026hyformer}, Mixformer~\cite{huang2026mixformer} and UniMixer~\cite{ha2026unimixer} organize heterogeneous information within a unified modeling framework and learn interactions across information sources for ranking.
    
\end{itemize}

In our evaluation, sequential interaction models are cascaded with feature interaction models to form complete ranking pipelines, while unified architectures are evaluated as integrated ranking models.

\subsubsection{\textbf{Implementation Details}.}

EASFormer is implemented in PyTorch and optimized at the operator level for the target deployment hardware.
We apply standard field-specific transformations $\phi_f(\cdot)$ to the input fields and concatenate the resulting representations before projection. The same encoding pipeline is used for EASFormer and all baselines to ensure a fair comparison, while detailed field definitions are omitted for confidentiality.

EASformer is optimized using AdamW with
$(\beta_1,\beta_2)=(0.9,0.98)$. We use a batch size of 64, learning rates of $1\times10^{-4}$ and $5\times10^{-5}$ for embedding and dense parameters, respectively, and a weight decay of $0.01$ for dense parameters. EASformer consists of $L=5$ blocks with 8 attention heads per block and uses $K=3$ NS tokens. The masked-action ratio and self-supervised loss weight are set to $\rho=0.2$ and $\lambda_s=1$, respectively.
Training and evaluation are conducted in FP32 on Ascend 910B3 NPUs.

We instantiate two variants of EASformer. Both use NS tokens with $\mathcal{D}_{\mathrm{NS}}=512$. The full EASformer sets the Environment, Action, and State representation dimension to $\mathcal{D}=512$, whereas the lightweight version EASformer$^{-}$ reduces $\mathcal{D}$ to 128. 
The maximum history length is set to $T=500$ for all offline experiments; for online deployment, we use $T=500$ for EASFormer$^{-}$ and $T=1{,}000$ for the production HyFormer baseline.



\subsection{Main Results}

\begin{table}[t]
    \centering
    \caption{Overall performance.}
    \label{tab:main_results}
    \small
    \renewcommand{\arraystretch}{1.08}

    \begin{tabular*}{0.88\columnwidth}{
        @{\extracolsep{\fill}}
        llrr
        @{}
    }
        \toprule
        \makecell[tl]{\textbf{Sequence}\\\textbf{Modeling}} &
        \makecell[tl]{\textbf{Feature}\\\textbf{Interaction}} &
        \makecell[tr]{\textbf{AUC}$\uparrow$} &
        \makecell[tr]{\textbf{$\Delta$AUC}\\\textbf{(\%)}} \\
        \midrule

        \multicolumn{4}{c}{\textbf{Two-Stage Models}} \\
        \midrule
        LONGER & RankMixer & 0.88895 & $+0.30$ \\
        LONGER & DCNv2     & 0.88660 & $+0.04$ \\
        LONGER & Wukong    & 0.88859 & $+0.26$ \\
        SASRec & RankMixer & 0.87410 & $-1.37$ \\
        SASRec & DCNv2     & 0.86540 & $-2.36$ \\
        SASRec & Wukong    & 0.87710 & $-1.04$ \\

        \midrule
        \multicolumn{4}{c}{\textbf{Unified Architectures}} \\
        \midrule
        \multicolumn{2}{@{}l}{OneTrans}
            & 0.88599 & $-0.03$ \\
        \multicolumn{2}{@{}l}{HyFormer (Baseline)}
            & 0.88628 & $0.00$ \\
        \multicolumn{2}{@{}l}{UniMixer}
            & 0.87984 & $-0.73$ \\
        \multicolumn{2}{@{}l}{MixFormer}
            & 0.88671 & $+0.05$ \\
        \multicolumn{2}{@{}l}{MTGR}
            & 0.88833 & $+0.23$ \\

        \midrule
        \multicolumn{2}{@{}l}{EASformer$^{-}$}
            & 0.89301 & $+0.76$ \\
        \multicolumn{2}{@{}l}{\textbf{EASformer (Ours)}}
            & \textbf{0.89411} & \textbf{$+0.88$} \\
        \bottomrule
    \end{tabular*}
\end{table}

We compare EASformer with representative baselines on a 25-day offline benchmark from a real-world industrial app marketplace, with results reported in Table~\ref{tab:main_results}. HyFormer, the deployed production model, serves as the reference for computing $\Delta$AUC, and the best result is highlighted in bold. We make the following observations.

\textbf{(i)} EASformer achieves the highest AUC of 0.89411, corresponding to relative improvements of 0.88\% over HyFormer, 0.58\% over the strongest two-stage baseline (LONGER+RankMixer), and 0.65\% over the strongest unified baseline (MTGR). These results support our semantic alignment design. Existing architectures may still directly interact sequence and NS representations with asymmetric semantics, whereas EASformer performs history--candidate matching in an aligned environment space, enabling more targeted retrieval of relevant historical feedback.

\textbf{(ii)} The comparison with MTGR further demonstrates the benefit of EASformer's structured interaction. Although our MTGR implementation also separates environment and action tokens, EASformer assigns them distinct roles: Environment representations determine history--candidate matching weights, Action representations provide the retrieved behavioral evidence, and the State stream complements direct retrieval. The improvement over MTGR indicates that explicit role-specific interaction provides benefits beyond input-level token separation.

\textbf{(iii)} We also report EASformer$^{-}$, a lightweight variant with 128-dimensional Environment, Action, and State representations, yielding an aggregate width of 384 across the three streams, compared with the 512-dimensional sequence representation in the baseline setting. Despite its reduced width, EASformer$^{-}$ achieves an AUC of 0.89301 and a 0.76\% relative improvement over HyFormer. This result suggests that the gain mainly arises from the aligned and disentangled modeling strategy rather than increased representation width, while also demonstrating its potential for lightweight online deployment.

\begin{table}[ht]
    \centering
    \small
    \setlength{\tabcolsep}{4pt}
    \caption{Ablation results. $\Delta$AUC denotes the relative change with respect to the full EASformer model.}
    \label{tab:ablation}
    \begin{tabular}{lrr}
        \toprule
        Variant & AUC & $\Delta$AUC (\%) \\
        \midrule
        EASformer (full)
            & \textbf{0.89411} & -- \\
        w/o SSL
            & 0.89146 & $-0.30$ \\
        w/o EAS
            & 0.88832 & $-0.65$ \\
        w/o Spatial Mapping
            & 0.89262 & $-0.17$ \\
        w/o Temporal Retrieval
            & 0.89084 & $-0.37$ \\
        \bottomrule
    \end{tabular}
\end{table}

\subsection{Ablation Study}
We evaluate four variants of EASformer: \textit{w/o SSL} removes masked-action self-supervision; 
\textit{w/o EAS} replaces the aligned and disentangled modeling with unified interaction over mixed historical and candidate representations;
\textit{w/o Spatial Mapping} retains only the Temporal Retrieval branch; and \textit{w/o Temporal Retrieval}  retains the Environment-stream self-attention and Spatial Mapping branch.
All relative AUC changes are measured against the full model.


Despite using the same input features, \textit{w/o EAS} causes the largest relative AUC drop of 0.65\%, supporting the importance of disentangling historical context from observed actions and aligning history--candidate semantics before interaction.


Removing either Temporal Retrieval or Spatial Mapping degrades performance, confirming the contribution of both views. The larger relative AUC drop without Temporal Retrieval indicates that retrieving historical actions from similar context provides more direct behavioral evidence, while state-conditioned action mapping contributes complementary signals by adapting prediction to the current user state. Combining the two views therefore achieves the best performance.

Removing masked-action self-supervision reduces AUC by 0.30\%, indicating that historical environment--action pairs provide useful auxiliary supervision. By treating masked interactions as pseudo-candidates and reusing the main prediction pathway, this objective remains aligned with the ranking task and improves performance.


\subsection{Qualitative Analysis of Semantic Alignment}

\begin{figure}[t]
    \centering
    \begin{subfigure}{\linewidth}
        \centering
        \includegraphics[width=\linewidth]
            {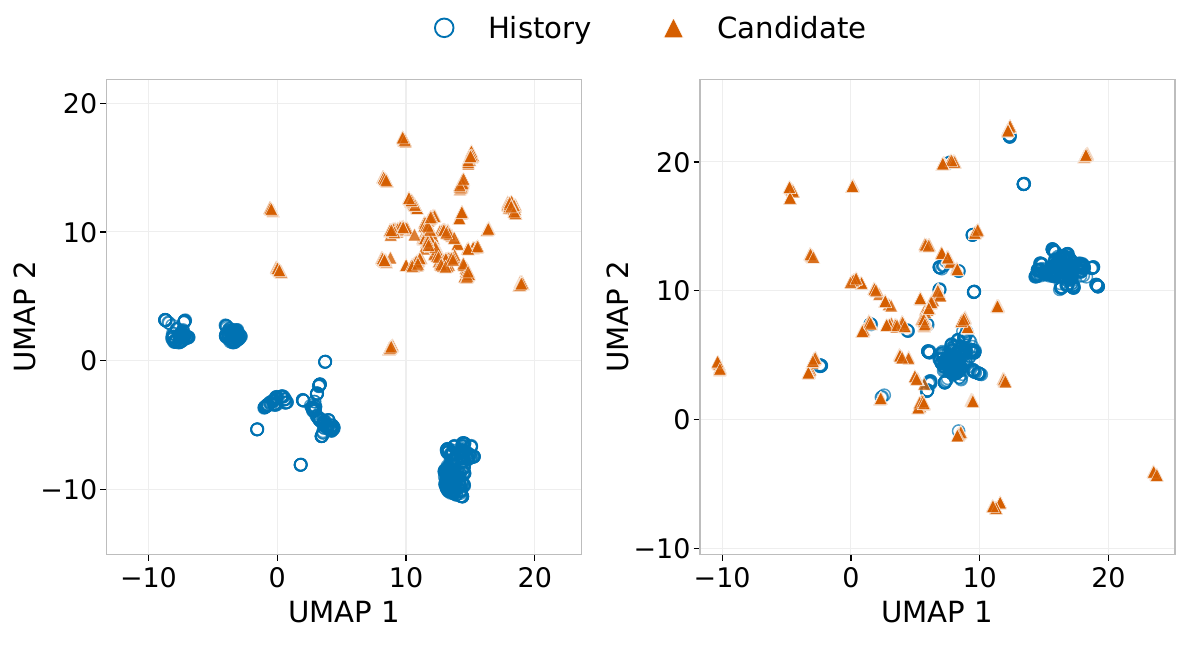}
        \caption{UMAP visualization
            (left: Unaligned; right: Aligned).}
        \label{fig:umap_visualization}
    \end{subfigure}

    \vspace{0.6em}

    \begin{subfigure}{\linewidth}
        \centering
        \includegraphics[width=\linewidth]
            {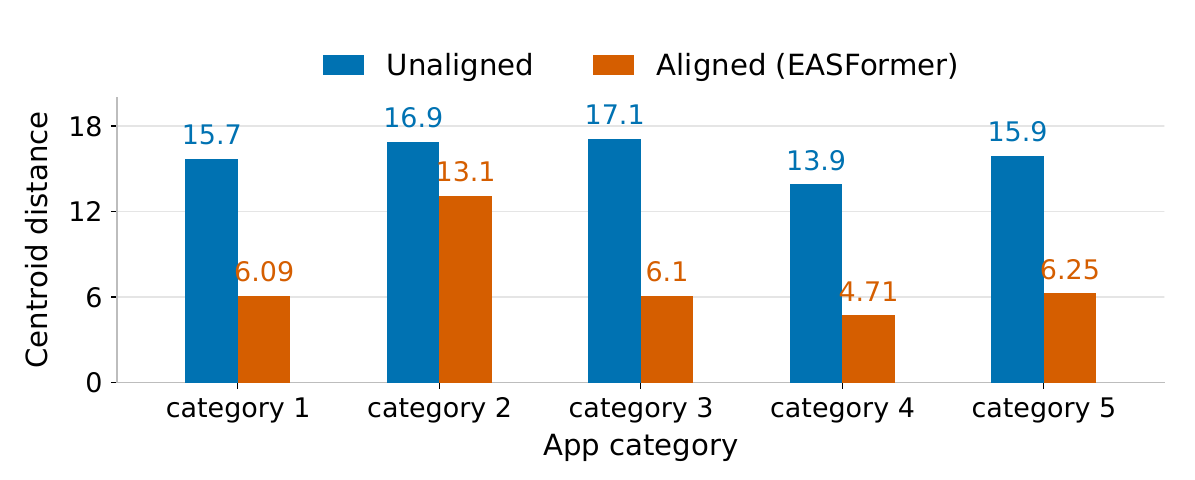}
        \caption{History--Candidate centroid distance in UMAP space.}
        \label{fig:centroid_distance}
    \end{subfigure}

    \caption{Qualitative comparison of history--candidate representation alignment.}
    \label{fig:semantic_alignment}
\end{figure}

\begin{figure}[ht]
    \centering
    \includegraphics[width=\linewidth]
    {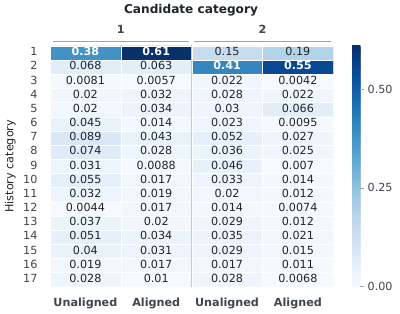}
    \caption{Category-level candidate-to-history attention. Bold headers denote candidate categories; row labels denote history categories.}
    \label{fig:ablation-attn}
\end{figure}

To examine environment–action disentanglement and semantic alignment, we compare EASformer (Aligned) with a model without environment–action field separation (Unaligned). We extract history and candidate representations from both models after tokenization, and project them into two dimensions using UMAP. Figure \ref{fig:umap_visualization} shows clear separation between history and candidate representations in Unaligned, whereas the two sources exhibit greater overlap in Aligned. We further compute separate history and candidate centroids within each app category and measure their Euclidean distance. As shown in Figure \ref{fig:centroid_distance}, Aligned exhibits smaller projected centroid distances across all displayed categories. These observations are consistent with EASformer’s design: environment–action disentanglement produces dedicated environment representations, while semantically corresponding fields and a shared encoder support history–candidate semantic alignment.

We further examine candidate-to-history attention using two illustrative candidate cases. For each candidate, we retain the attention weights assigned to valid and visible historical interactions and compute the mean attention within each app category. As shown in Figure \ref{fig:ablation-attn}, Aligned assigns higher normalized mean attention than Unaligned to historical interactions in the candidate’s own category in both cases. This pattern is consistent with EASformer’s interaction mechanism, in which environment representations determine history–candidate matching weights that guide the aggregation of historical action representations.

\begin{figure}[ht]
    \centering
    \includegraphics[width=\linewidth]
    {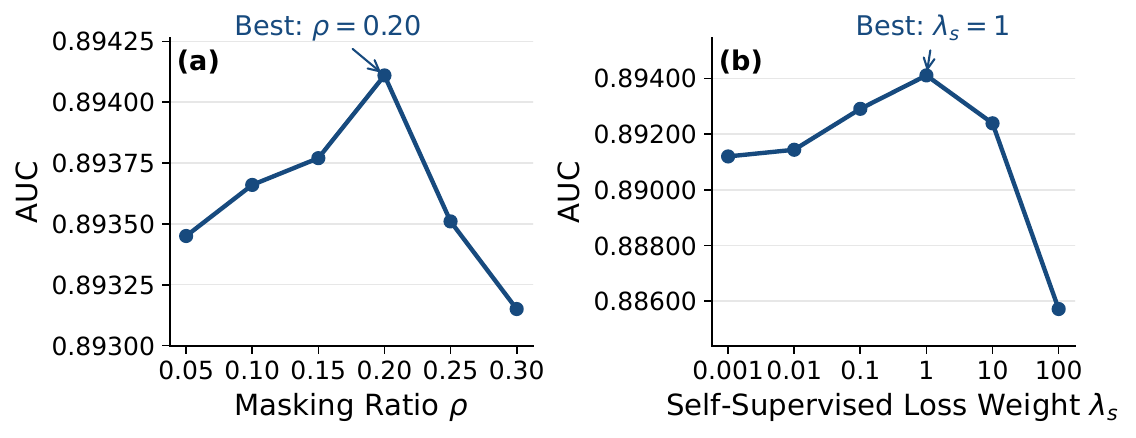}
    \caption{Parameter sensitivity of the masked-action self-supervised
    objective. (a) Masking ratio $\rho$. (b) Self-supervised loss
    weight $\lambda_s$.}
    \label{fig:parameter-sensitivity}
\end{figure}

\subsection{Parameter sensitivity analyses}
We conduct a hyperparameter sensitivity analysis on the masking ratio $\rho$ and the self-supervised loss weight $\lambda_s$. As shown in Figure~\ref{fig:parameter-sensitivity}, EASformer achieves the best performance at $\rho=0.20$ and $\lambda_s=1$. A small masking ratio provides insufficient self-supervised signals, whereas an excessively large ratio masks too much historical action information and weakens history--candidate interaction. Similarly, a small $\lambda_s$ underutilizes the auxiliary objective, while an overly large value causes it to dominate the primary ranking task. We therefore set $\rho=0.20$ and $\lambda_s=1$ in the following experiments.

\subsection{Efficiency Analysis}
\label{sec:efficiency}

We analyze the effects of history length and hidden dimension on the efficiency--effectiveness trade-off of EASformer. Table~\ref{tab:efficiency} reports AUC and model-only inference latency under the same serving environment. The latency is averaged over 10 inference samples.

\begin{table}[ht]
    \centering
    \caption{Efficiency analysis of EASformer.
    $\mathcal{D}$ denotes the hidden dimension of each Environment, Action, and State stream, and
    $\mathcal{D}_{\mathrm{NS}}=512$ for all settings.}
    \label{tab:efficiency}

    \small
    \renewcommand{\arraystretch}{1.08}

    \setlength{\arrayrulewidth}{0.4pt}

\begin{tabular}{
        @{}
        l
        c
        |
        S[table-format=1.5]
        S[table-format=2.2]
        @{}
    }
        \toprule
        $\boldsymbol{\mathcal{D}}$ &
        \textbf{History Length ($T$)} &
        \multicolumn{1}{c}{\textbf{AUC}$\uparrow$} &
        \multicolumn{1}{c}{\textbf{Latency (ms)}$\downarrow$} \\
        \midrule

        512 & 500 & \bfseries 0.89411 & 39.40 \\
        512 & 375 & 0.89345 & 39.23 \\
        512 & 250 & 0.89308 & 37.20 \\
        512 & 125 & 0.89135 & 35.14 \\
        128 & 500 & 0.89301 & \bfseries 27.71 \\

        \bottomrule
\end{tabular}
\end{table}

Shortening the history provides moderate latency savings at the cost of ranking accuracy. For example, reducing the history length from 500 to 250 lowers latency by 5.6\% with a 0.12\% relative AUC decrease, while a length of 125 produces a larger accuracy drop. 
Reducing $\mathcal{D}$ to 128 is more effective: EASformer$^{-}$ decreases latency by 29.7\% while retaining an AUC of 0.89301, only 0.12\% lower than the full model. We therefore adopt EASformer$^{-}$ with a history length of 500 for online deployment.

\subsection{Online Evaluation}

Following the offline evaluation, we deploy EASformer$^{-}$ in two real-world industrial app-ranking scenarios and conduct online A/B tests against the production baseline model (HyFormer) over one evaluation period. 
Due to commercial confidentiality, the scenario names and absolute metric values are anonymized, and only relative eCPM lifts are reported.

\begin{table}[ht]
    \centering
    \caption{Relative eCPM lifts of EASformer$^{-}$ over the production HyFormer baseline during one online evaluation period.}
    \label{tab:online_results}
    \small
    \renewcommand{\arraystretch}{1.08}
    \setlength{\tabcolsep}{12pt}

    \begin{tabular}{@{}lr@{}}
        \toprule
        \textbf{Scenario} & \textbf{eCPM Lift}$\uparrow$ \\
        \midrule
        Ranking Scenario 1 & $+1.26\%$ \\
        Ranking Scenario 2 & $+3.94\%$ \\
        \bottomrule
    \end{tabular}
\end{table}

As shown in Table~\ref{tab:online_results}, EASformer$^{-}$ achieves positive eCPM lifts of 1.26\% and 3.94\% in the two ranking scenarios, respectively. These results demonstrate that its offline improvements can translate into practical gains in real-world online serving.

We further report model-only latency. Under the deployed configurations, EASformer$^{-}$ uses a history length of 500 and achieves an average latency of 27.71 ms, whereas HyFormer uses a history length of 1,000 and requires 33.68 ms. At the same history length of 500, HyFormer is faster at 21.11 ms. These results show that EASformer$^{-}$ provides a practical quality--latency trade-off: despite using a shorter history, it satisfies online latency requirements while delivering the positive performance gains reported above.


\section{Related Work}
We review prior work most relevant to EASformer as follows.


\subsection{Sequential and Generative Recommendation}

Wide \& Deep combines memorization and generalization; DIN and DIEN model candidate-conditioned interests and temporal interest evolution, respectively~\cite{cheng2016wide,zhou2018din,zhou2019dien}. TWIN V2 handles ultra-long histories through hierarchical clustering and cluster-aware target attention~\cite{si2024twinv2}. SASRec and BERT4Rec use self-attentive and bidirectional modeling, respectively; TransAct models real-time action histories with Transformers~\cite{kang2018sasrec,sun2019bert4rec,xia2023transact}.

Conventional rankers often separate history encoding from feature interaction and reprocess shared histories~\cite{zhang2025onetrans,zhai2024hstu,han2025mtgr}. HSTU reformulates retrieval and ranking as generative sequential transduction with history reuse and empirical power-law scaling with training compute~\cite{zhai2024hstu}. MTGR adapts HSTU with candidate-side cross features and discriminative training~\cite{han2025mtgr}. OneTrans jointly models sequential and non-sequential (NS) tokens within Transformer blocks~\cite{zhang2025onetrans}.

\subsection{Heterogeneous Feature Interaction}

Factorization Machines model pairwise interactions; DeepFM adds neural feature learning~\cite{rendle2010fm,guo2017deepfm}. xDeepFM and DCNv2 learn explicit crosses, while FCN fuses linear and exponential cross networks~\cite{lian2018xdeepfm,wang2021dcnv2,li2026fcn}. AutoInt uses self-attention, CAN conditions network parameters on input features, and Wukong scales stacked interaction blocks~\cite{song2018autoint,bian2020can,wang2024wukong}; SFG regenerates features under click supervision~\cite{yin2025sfg}.

Feature heterogeneity also shapes interaction design~\cite{zhu2025rankmixer,huang2026mixformer}. RankMixer uses parameter-free multi-head token mixing with token-specific FFNs~\cite{zhu2025rankmixer}; MTGR uses group-layer normalization for heterogeneous token groups~\cite{han2025mtgr}. InterFormer exchanges selective summaries between sequence and non-sequence branches, whereas OneTrans uses a unified Transformer~\cite{zeng2025interformer,zhang2025onetrans}. HyFormer and MixFormer couple query mixing with cross-attention~\cite{huang2026hyformer,huang2026mixformer}; HeMix extracts interests using dynamic and learned fixed queries before heterogeneous mixing~\cite{wang2026hemix}. EST combines lightweight cross-attention with content-guided sparse intra-sequence interactions~\cite{liu2026est}.

EASformer uses shared encoders for corresponding historical and candidate Environment fields, separates Actions, and reserves a fusion pathway for supplementary NS context. Environment-derived attention retrieves Action evidence.

\subsection{Historical Auxiliary Supervision}

To augment CTR training, DIEN discriminates next clicked items from negatives, MISS adds interest-level contrastive supervision, and AT4CTR adds user--item matching and next-item prediction~\cite{zhou2019dien,guo2022miss,liu2024at4ctr}. SRP4CTR transfers bidirectional masked-item and masked-behavior pretraining to CTR~\cite{han2024srp4ctr}. UniPinRec masks historical actions while retaining item information and predicts historical and candidate responses with shared action-specific heads under causal history attention~\cite{li2026unipinrec}. EASformer jointly trains candidate ranking and auxiliary masked historical action reconstruction, retaining the corresponding Environment fields.



\section{Conclusion}

In this work, we identify semantic misalignment between historical sequence representations and candidate-side non-sequential features as a key challenge in industrial generative ranking. We propose EASformer, which constructs an aligned context-environment space while disentangling observed actions and supplementary NS information.

Building on this design, EASformer combines environment-guided temporal retrieval with state-conditioned spatial mapping, and further exploits historical environment--action relations through masked-action self-supervision. 

Experiments on a real-world industrial app-ranking benchmark show that EASformer achieves an AUC of 0.89411, corresponding to a 0.88\% relative improvement over the deployed baseline. The lightweight EASformer$^{-}$ provides a practical efficiency--effectiveness trade-off, while online A/B tests yield relative eCPM lifts of 1.26\% and 3.94\% across two real-world ranking scenarios. These results demonstrate the practical value of semantic alignment, role-specific interaction, and historical auxiliary supervision for industrial recommendation.

\bibliographystyle{ACM-Reference-Format}
\bibliography{references}

\end{document}